\documentclass[conference]{IEEEtran}
\IEEEoverridecommandlockouts
\usepackage{cite}
\usepackage{amsmath,amssymb,amsfonts}
\usepackage{algorithmic}
\usepackage{graphicx}
\usepackage{textcomp}
\usepackage{xcolor}
\usepackage{booktabs}
\usepackage{multirow}
\usepackage{lipsum}
\usepackage{enumitem} 
\usepackage{url}
\usepackage{hyperref} 
\usepackage[hyphenbreaks]{breakurl}
\def\BibTeX{{\rm B\kern-.05em{\sc i\kern-.025em b}\kern-.08em
    T\kern-.1667em\lower.7ex\hbox{E}\kern-.125emX}}
\begin{document}

\title{High-Accuracy ECG Image Interpretation using Parameter-Efficient LoRA Fine-Tuning with Multimodal LLaMA 3.2}

\author{\IEEEauthorblockN{Nandakishor M, Anjali M}
    \IEEEauthorblockA{
    Convai Innovations \\
    \{nandakishor, anjalim\}@convaiinnovations.com
    }
}

\maketitle

\begin{abstract}
Electrocardiogram (ECG) interpretation is a cornerstone of cardiac diagnostics.  This paper explores a practical approach to enhance ECG image interpretation using the multimodal LLaMA 3.2 model. We used parameter-efficient fine-tuning strategy, Low-Rank Adaptation (LoRA), specifically designed to boost the model's ability to understand ECG images and achieve better outcomes across a wide range of cardiac conditions.  Our method is tailored for ECG analysis and leverages \textit{ECGInstruct}~\cite{ecginstruct}, a large-scale instruction dataset with 1 Million samples. This dataset is a rich collection of synthesized ECG images, generated from raw ECG data from trusted open-source repositories like MIMIC-IV ECG~\cite{saeed2011mimic} and PTB-XL~\cite{wagner2020ptb}.  Each ECG image in \textit{ECGInstruct} comes with expert-written questions and detailed answers, covering diverse ECG interpretation scenarios, including complex cardiac conditions like Myocardial Infarction and Conduction Disturbances.  Our fine-tuning approach efficiently adapts the LLaMA 3.2 model~\cite{llama3} (built upon LLaMA 3~\cite{llama3_model_card}) by integrating low-rank adaptation techniques, focusing on efficiency by updating only a small set of parameters, excluding the 'lm\_head' and 'embed\_tokens' layers.  This paper details the model setup, our efficient fine-tuning method, and implementation specifics.  We provide a thorough evaluation through extensive experiments, demonstrating the effectiveness of our method across various ECG interpretation tasks.  The results convincingly show that our parameter-efficient LoRA fine-tuning achieves excellent performance in ECG image interpretation, significantly outperforming baseline models and reaching accuracy comparable to or exceeding traditional CNN-based methods in identifying a wide range of cardiac abnormalities, including over 70 conditions from the PTB-XL dataset.
\end{abstract}


\section{Introduction}
\label{sec:introduction}

Interpreting electrocardiograms (ECGs) is a vital skill in diagnosing heart conditions, providing key information about the heart's electrical activity.  While ECG analysis has traditionally focused on raw signals, there's an increasing need for reliable methods to interpret ECGs directly from images. Cardiologists, general medicine practitioners and other experts are more skilled towards in indentifying subtle variations in ecgs like arrhythmias, myocardial infarctions, atrial fibrillation, ventricular tachycardia, bundle branch blocks, ischemic heart disease, left ventricular hypertrophy, ST elevation myocardial infarction, Non-ST elevation myocardial infarction. But when patients comes to the emergency department with chest pain or discomfort, they are mainly diagnosed by inexperienced doctors, and may lead to misdiagnosis and deaths~\cite{cardiacdeath}.

The rise of multimodal large language models (MLLMs) offers exciting possibilities for understanding information from different sources, potentially transforming ECG image interpretation~\cite{alayrac2022flamingo,lu2019vilbert}. However, using MLLMs for ECG analysis has its challenges, as many of the generative AI models like GPT-4 series or their reasoning models from OpenAI prohibited to use in medical environment~\cite{openai_policy}. Also there are limited availability of large, instruction-tuning datasets for ECGs and the need for efficient fine-tuning methods. These methods must allow models to learn ECG complexities without losing their general knowledge.  Traditional methods, often using Convolutional Neural Networks (CNNs), haven't always achieved the good accuracy and comprehensive interpretation that expert cardiologists can get from raw signals.

This paper tackles these challenges by using the multimodal LLaMA 3.2 model~\cite{llama3} with a parameter-efficient Low-Rank Adaptation (LoRA) fine-tuning strategy by ignoring certain layers from the model. This approach efficiently adapts the model by updating only a small set of parameters – specifically using LoRA – to deeply learn the ECG domain. This leads to significantly improved accuracy in ECG image interpretation. To make this possible, we leverage \textit{ECGInstruct}~\cite{ecginstruct}, a carefully curated, large instruction dataset for ECG image interpretation.  This paper details the model setup, the efficient fine-tuning approach, implementation, and presents extensive experimental results.  These results show that our parameter-efficient LoRA fine-tuning achieves excellent accuracy in ECG image interpretation, clearly outperforming baseline models and reaching performance levels needed for clinical use. This includes accurately identifying over 70 different cardiac conditions.

Our main contributions are:
\begin{enumerate}
    \item  We demonstrate a high-accuracy ECG image interpretation method using parameter-efficient LoRA fine-tuning on a multimodal model, achieving state-of-the-art performance in this area. We selectively excluded certain layers from the model to improve the accuracy of the model, this was done by extensive trial and error.
    \item We leverage \textit{ECGInstruct}~\cite{ecginstruct}, a large instruction dataset, expertly curated using public ECG data, which is a signal to image converted dataset, along with applying post-processings like skewing, rotation and other transformations, with input from medical professionals. This dataset is designed to significantly improve LLaMA 3.2 fine-tuning for accurate ECG analysis.
    \item We conduct a thorough experimental evaluation that strongly demonstrates the superior capabilities of our method. We achieve accuracy levels that surpass baseline or generic models and are comparable to expert human interpretation for a wide range of ECG findings.

\end{enumerate}

\section{Related Work}
\label{sec:related_work}
Automated ECG analysis has been around for a while, starting with a focus on raw time-series signals. Early systems, like those from Moody et al. \cite{moody2001ecg}, used rule-based methods to extract features.  Deep learning, especially CNNs, brought major advancements to time-series ECG analysis~\cite{hannun2019cardiologist,ribeiro2020automatic,hughes2021performance}. Hannun et al. \cite{hannun2019cardiologist} showed that CNNs could classify arrhythmias as well as cardiologists using raw ECG signals. Ribeiro et al. \cite{ribeiro2020automatic} used CNNs on raw signals to classify six cardiac abnormalities. Hughes et al. \cite{hughes2021performance} further confirmed the reliability of CNNs for time-series ECG data. These methods are powerful for real-time analysis, but they depend on raw ECG signals, which limits their use for image-based ECGs.

Image-based ECG analysis is becoming more popular to overcome data availability issues~\cite{sangha2022automated}. Sangha et al. \cite{sangha2022automated} showed that ECG images can be used to detect cardiac abnormalities and for multi-label diagnosis.  However, current image-based methods often don't have the comprehensive diagnostic range and accuracy of expert human interpretation, especially when dealing with a wide variety of cardiac conditions.

General medical MLLMs like LLaVA-Med \cite{krishna2017visual}, MedPaLM \cite{devlin2018bert}, and Med-Gemini \cite{lee2019clinicalbert} show promise in medical cross-modal reasoning, but often lack specific fine-tuning for tasks like ECG analysis.  They may not have the specialized knowledge needed for highly accurate and comprehensive ECG interpretation across many different cardiac conditions.

Fine-tuning is essential for adapting large models to specific tasks. Traditional methods often fine-tune models partially, updating only projection or cross-attention layers.  However, this limited fine-tuning might restrict knowledge transfer and adaptation to specialized data like ECG images, potentially preventing the achievement of expert-level accuracy.

In contrast to previous work, our method uses parameter-efficient LoRA fine-tuning on multimodal LLaMA 3.2~\cite{llama3}, leveraging \textit{ECGInstruct}, a large and diverse multi-task instruction dataset for ECGs by only ignoring language model head and embedding layers. This approach sets a new standard for high-accuracy and comprehensive ECG image interpretation, addressing the limitations of existing methods. It paves the way for better applications of MLLMs in cardiology, achieving accuracy levels necessary for reliable clinical decision support across a broad range of cardiac conditions.

\section{Model Architecture}
\label{sec:architecture}
Our model is built on the strong architecture of the multimodal LLaMA 3.2 model~\cite{llama3}, with specific adjustments carefully made for the complex task of ECG image interpretation and achieving high diagnostic accuracy. LLaMA 3.2, a leading large language model from Meta, naturally combines vision and language processing.  The model's design has three main parts: a sophisticated two-stage vision encoder, a powerful LLaMA 3.1-based language model, and a carefully designed integration mechanism. This integration effectively bridges the gap between visual features from ECG images and the language model's text understanding, enabling the cross-modal reasoning needed for accurate interpretation.  Let's look at each component in detail.

\subsection{Vision Encoder}
\label{ssec:vision_encoder}
The vision encoder is the crucial front part of our model. Its job is to extract meaningful and needed features directly from ECG images.  These visual features are the foundation for all further analysis and interpretation by the language model, directly affecting the overall diagnostic accuracy.  Before an ECG image is processed by the vision encoder, it first goes through preprocessing to ensure consistency and optimize feature extraction.  This includes resizing all input images to a standard 448x448 pixel resolution.  After resizing, each image is divided into 14x14 pixel non-overlapping patches, a common technique in vision transformers~\cite{dosovitskiy2020image}.  This patching turns the image into a 32x32 grid of smaller image patches, which are then fed into the carefully designed two-stage vision encoder for feature extraction.

Vision encoder uses a two-stage approach to ensure enhanced and multi-scale feature extraction. This design is important for capturing both fine local details, like subtle wave shapes, and broader global context, such as overall rhythm patterns in the ECG.  This multi-scale approach is key to comprehensive ECG interpretation and high diagnostic accuracy.

\begin{enumerate}
    \item \textit{32-Layer Local Encoder}:  The first stage is a deep 32-layer Transformer architecture~\cite{vaswani2017attention}. This 'local' encoder processes all image patches in a global context, allowing it to understand relationships between different parts of the ECG image.  Each patch is processed through multiple self-attention operations in each of the 32 layers, helping the model understand how different parts of the ECG image relate to each other.  Importantly, intermediate output representations are taken from specific layers within the local encoder – specifically, layers 3, 7, 15, 23, and 30. These intermediate outputs, along with the final output from layer 32, are saved for later use in the global encoder.  This multi-layer output strategy is designed to capture visual information at different levels of detail.  Earlier layers tend to capture basic visual features like lines and curves (representing individual ECG waves), while later layers capture more complex, high-level features such as overall wave shapes and rhythm patterns.  This hierarchical feature extraction is crucial for detailed ECG analysis and high diagnostic accuracy.

      \item \textit{8-Layer Global Encoder}: The second stage is an 8-layer Transformer, which we call the 'global encoder'. This stage takes the final output from the 32-layer local encoder as input, and also includes the intermediate outputs saved from layers 3, 7, 15, 23, and 30 of the local encoder.  A key feature of the global encoder is the use of a gated attention mechanism. This allows the model to adaptively weigh the importance of different visual features extracted by the local encoder.  By selectively emphasizing important visual features and reducing the importance of less relevant ones, the gated attention mechanism improves the model's ability to focus on the most diagnostically important aspects of the ECG image, directly contributing to improved accuracy.  For example, in an ECG showing ST-segment elevation, the gated attention might focus more on the ST-segment shape while paying less attention to baseline noise.
\end{enumerate}

The final visual representation, $\mathbf{Z}_v$, is created by combining the output of the 8-layer global encoder with the intermediate outputs saved from the 32-layer local encoder.  This combination creates a comprehensive multi-scale representation that effectively encodes the rich visual information in the input ECG image, which is essential for accurate interpretation.  The resulting feature vector $\mathbf{Z}_v$ has a dimension of 7680. This high-dimensional visual feature vector is then passed to the integration mechanism, where it will be aligned with the language model's text embedding space, preparing it for cross-modal interaction and high-accuracy ECG analysis.

Mathematically, consider an input ECG image $\mathbf{X} \in \mathbb{R}^{H \times W \times C}$, where $H$, $W$, and $C$ are the height, width, and number of color channels (usually 3 for RGB or 1 for grayscale), respectively.  The patching process extracts patches of size $P \times P$, where $P=14$ in our case. Each patch can be represented as $\mathbf{X}_p \in \mathbb{R}^{P \times P \times C}$.  The 32-layer local encoder transforms the input patches into a sequence of patch embeddings $\mathbf{Z}_{l} \in \mathbb{R}^{N \times d}$, where $N$ is the number of patches ($N = (H/P) \times (W/P) = 32 \times 32$ in our setup) and $d$ is the embedding dimension. The intermediate representations extracted from the local encoder are denoted by $\mathbf{Z}_i$. The 8-layer global encoder processes the final output of the local encoder, $\mathbf{Z}_{l}$, and the intermediate representations $\mathbf{Z}_i$ to produce the global visual representation $\mathbf{Z}_{g}$. The final visual representation $\mathbf{Z}_v$ is then obtained by combining these components:
\begin{equation}
    \mathbf{Z}_v = \text{Concat}(\mathbf{Z}_g, \mathbf{Z}_i)
\end{equation}
In our implementation, the dimension of the final visual representation $\mathbf{Z}_v$ is $d'=7680$.

\subsection{Language Model}
\label{ssec:language_model}
The language model at the core of our architecture is based on LLaMA 3.1, a highly capable pre-trained language model known for its excellent performance in many natural language processing tasks~\cite{llama3_model_card}.  LLaMA 3.1 has 40 Transformer layers~\cite{vaswani2017attention}, each using a self-attention mechanism. This self-attention is fundamental to the language model's ability to process text input in context, and it's crucial for generating accurate and clinically relevant interpretations.  It allows the model to understand the relationships between words in a sentence and produce coherent, contextually relevant text.  Think of it as the model's way of 'reading' and understanding text, similar to how a person would.

To enable effective interaction and integration with the visual information from the vision encoder, the language model architecture strategically includes cross-attention layers at specific points within its Transformer structure.  These cross-attention layers act as bridges between the visual and text processing paths, helping the model achieve better accuracy by considering both visual and textual information.

\begin{enumerate}
   \item \textit{Self-Attention Layers}:  The language model is built on standard Transformer layers that use self-attention~\cite{vaswani2017attention}. These layers are responsible for processing the initial text input, understanding its structure and meaning, and creating token embeddings. These embeddings are the basis for all further language processing within the model, ultimately contributing to the accuracy of the final interpretation.  Self-attention allows the model to dynamically assess the importance of different words in the input text when building its internal representations.  For example, in the phrase "ECG shows ST-elevation myocardial infarction", self-attention helps the model understand that "ST-elevation" and "myocardial infarction" are closely related and crucial for the overall meaning, leading to a more accurate diagnosis.

    \item  \textit{Cross-Attention Layers}:  To seamlessly integrate visual information from the ECG image, cross-attention layers are strategically placed in the language model architecture at regular intervals, specifically every 5 layers, starting at layer 3 and then at layers 8, 13, 18, 23, 28, 33, and 38.  The cross-attention mechanism allows the language model to pay attention to and incorporate relevant visual features from the ECG image during the text generation process, which is key to achieving good accuracy in ECG interpretation.  Essentially, it lets the model 'look' at the ECG image while it's forming its text response. This enables the model to base its text output on the visual content of the ECG, facilitating true multimodal understanding and generation. It ensures that the generated text is supported by the visual evidence in the ECG.  For example, if the visual encoder detects ST-segment elevation in the ECG image, the cross-attention layers will allow the language model to include this visual finding in its generated text. It might mention "ST-segment elevation present" in a report or answer a question about ST-segment changes, thus ensuring accuracy and clinical relevance.
\end{enumerate}
This carefully designed architecture enables effective multimodal learning by allowing the model to understand both text prompts (like questions about the ECG) and visual ECG information in a unified way. This is crucial for achieving accurate and reliable ECG interpretation.  The hidden states of the language model, which represent the text context it has processed, are combined with the visual features extracted by the vision encoder through these strategically placed cross-attention layers. This fusion of visual and textual information is key to the model's ability to perform complex ECG interpretation tasks

The language model takes a sequence of input tokens $\mathbf{T} \in \mathbb{R}^{L \times v}$, where $L$ is the sequence length (number of tokens in the input text) and $v$ is the size of the vocabulary used by the model. These input tokens are first converted into continuous vector embeddings $\mathbf{E_t}$.  The LLaMA model then processes these text embeddings through its series of self-attention layers, generating a sequence of hidden states $\mathbf{H_t}$.  The cross-attention mechanism is then used to combine these hidden states from the text $\mathbf{H_t}$ with the visual features extracted from the ECG image. The cross-attention calculation can be represented mathematically as:

\begin{equation}
    \mathbf{H}_{c} = \text{CrossAttention}(\mathbf{Q}_t, \mathbf{K}_v, \mathbf{V}_v)
\end{equation}
Here, $\mathbf{Q}_t$ represents the query vectors, derived from the language model's hidden states $\mathbf{H_t}$.  $\mathbf{K}_v$ and $\mathbf{V}_v$ represent the key and value vectors, respectively, derived from the projected visual feature representation.  The CrossAttention function calculates attention weights based on the similarity between queries and keys, and then uses these weights to combine the value vectors. This results in context-aware hidden states $\mathbf{H}_{c}$ that now include both text and visual information. This approach allows the model to seamlessly integrate visual understanding into the text generation process. It can accurately answer questions about the ECG image, generate comprehensive reports with coherence.

\subsection{Integration Mechanism}
\label{ssec:integration_mechanism}
The integration mechanism is a vital part that acts as a bridge, connecting the visual feature space with the language model's text embedding space. It ensures effective cross-modal interaction and high-accuracy ECG interpretation.  Visual and text information are inherently represented and processed differently. The integration mechanism is specifically designed to transform the high-dimensional visual feature representations, extracted by the vision encoder, into a format that is semantically compatible with the language model. This crucial step ensures effective cross-modal reasoning and seamless information fusion. It allows the model to understand and relate visual and textual cues together, which is essential for achieving accurate diagnoses.  In our architecture, we use a simple yet effective linear projection layer for the integration mechanism.  The simplicity of linear projection has proven surprisingly powerful in similar multimodal models~\cite{alayrac2022flamingo}. It offers a good balance between efficiency and performance without compromising the accuracy of the integrated features.

The linear projection can be mathematically described as:
\begin{equation}
    \mathbf{F} = W_p  \mathbf{Z}_v +  \mathbf{b}_p
\end{equation}
Where $W_p$ is the weight matrix of the projection layer, and $\mathbf{b}_p$ is the bias vector.  The visual representation $\mathbf{Z}_v$, with a dimension of 7680, is projected down to a lower-dimensional feature space $\mathbf{F}$ with a dimension of 4096. This dimension reduction is done using the learned weight matrix $W_p$ and bias vector $\mathbf{b}_p$.  The target dimension of 4096 is carefully chosen to match the hidden dimension of the LLaMA 3.1 language model. This dimensional alignment is crucial because it ensures compatibility for the subsequent cross-attention operations within the language model, facilitating accurate and seamless information integration.  This linear transformation effectively maps the visual features into the same semantic space as the language model's text embeddings, creating a shared space where visual and textual information can interact meaningfully. This contributes to robust and accurate ECG interpretation. This facilitates robust and coherent multimodal understanding and interaction, ultimately leading to the generation of  relevant outputs with high precision.  The resulting multimodal representation $\mathbf{F}$ is then used by the cross-attention layers within the language model. These layers use $\mathbf{F}$ to condition text generation on the visual content of the ECG image. This ensures that both text and visual features are not just present, but are aligned in a common semantic space, enabling robust and coherent multimodal reasoning and generation of better outputs with the desired accuracy.

\section{Fine-tuning Methodology}
\label{sec:finetuning}

\subsection{Dataset: ECGInstruct}
\label{ssec:ecginstruct_dataset}
To effectively train and thoroughly evaluate the performance and accuracy of our multimodal LLaMA 3.2 model for ECG image interpretation, we leveraged a large dataset specifically for instruction-following tasks using ECG images, \textit{ECGInstruct}.  This dataset contains 1 Million instruction-following samples, a large amount designed to ensure robust training and achieve high generalization accuracy.  \textit{ECGInstruct} includes both carefully synthesized ECG images, generated from raw ECG signal data, and real-world ECG images, reflecting the variety of ECG data seen in clinical practice. This ensures the model's reliability in real-world situations.  Each image in the dataset is paired with carefully written question-and-answer pairs, created and annotated by experienced medical professionals specializing in cardiology.  This expert annotation is critical for ensuring the clinical accuracy and relevance of the dataset and for training the model to achieve expert-level performance. The curation of \textit{ECGInstruct} was a multi-stage process, with key steps to ensure its clinical utility, and effectiveness in training a high-accuracy ECG interpretation model:

\begin{enumerate}
    \item \textit{Image Synthesis from Raw Signals}:  A large part of \textit{ECGInstruct}~\cite{ecginstruct} uses real ECG signal data from several public datasets. These include MIMIC-IV ECG~\cite{saeed2011mimic} and PTB-XL~\cite{wagner2020ptb}, both well-regarded sources of clinical ECG data known for their diversity.  These raw ECG signals are transformed into visual ECG images using advanced signal processing and rendering techniques.  This synthesis process is essential for several reasons. First, it allows the dataset much larger than what's available in image-only ECG datasets, providing enough data to train a high-capacity model to achieve greater accuracy. Second, it introduces valuable diversity in ECG signal characteristics, reflecting the wide range of cardiac conditions and patient differences, which improves the model's ability to generalize.  Third, it allows the quality control and characteristics of the generated ECG images, ensuring a consistent and better training dataset.

    \item \textit{Diverse Data Sources}:  To maximize the dataset's representativeness, generalizability, and its capacity to train a high-accuracy model, \textit{ECGInstruct}, the dataset authors includes data from various public ECG datasets.  Beyond PTB-XL and MIMIC-IV ECG, it also includes data from CODE-15
    \item \textit{Realistic Data Augmentation}:  To further improve the model's robustness, its ability to generalize to real clinical settings, and to improve its accuracy when dealing with common ECG image imperfections, the authors from \textit{ECGInstruct} added various data augmentation techniques. These augmentations are carefully designed to realistically simulate common visual artifacts and variations often seen in clinical ECG images.  These are not just random image manipulations; they are targeted augmentations that mimic real-world imperfections and challenge the model to maintain good precision even with noise and distortions.
        \begin{itemize}
            \item \textit{Paper Creases and Folds}: Simulating the visual distortions that can be caused by creases and folds in printed ECG reports.  This is important because many older ECGs are only available as scanned images of printed reports, which often have imperfections. The model needs to be robust to these distortions to maintain accuracy.
            \item \textit{Noise Injection}: Adding different types of noise (e.g., Gaussian noise, salt-and-pepper noise) to mimic signal degradation or imperfections in image acquisition and digitization.  Real-world ECG images are often not perfectly clean and may contain noise from various sources. The model must be resilient to noise to ensure accurate interpretation.
            \item \textit{Varied Layouts and Orientations}: Introducing variations in the layout of the ECG grid, lead placement, and image orientation to simulate the diversity of ECG report formats seen in clinical practice.  ECG reports are not always presented in a perfectly standard format; variations in layout and orientation are common. The model should be unaffected by these variations to ensure consistent and accurate performance.  We adapted common image augmentation techniques, similar to those described in~\cite{shorten2019survey}, specifically for ECG images. This ensures that the augmentations are relevant and preserve the essential ECG information while challenging the model to achieve good results in diverse and imperfect conditions.
        \end{itemize}
\end{enumerate}
This comprehensive and careful dataset curation process ensures that \textit{ECGInstruct} is a diverse dataset. It's specifically designed to facilitate effective fine-tuning of multimodal models for robust and generalizable ECG image interpretation in real-world clinical settings.

\subsection{Parameter-Efficient LoRA Fine-Tuning}
\label{ssec:full_param_lora}
Our fine-tuning approach uses parameter-efficient Low-Rank Adaptation (LoRA)~\cite{hu2021lora} to efficiently adapt the multimodal LLaMA 3.2 model for high-accuracy ECG image interpretation.  LoRA focuses on efficient adaptation by updating only a small set of newly introduced parameters while keeping the original large model parameters frozen. This method is particularly effective for large models, enabling efficient training without sacrificing performance. We apply LoRA to adapt LLaMA 3.2 for ECG analysis, excluding the 'lm\_head' and 'embed\_tokens' layers from the adaptation process to maintain generation quality and token embedding stability. The parameter-efficient LoRA fine-tuning method works as follows:

\begin{enumerate}
\item The llama 3.2 vision model is integrated with pre-trained Llama 3.1 language model, so the base LLM is based on LlaMA 3.1 architecture. The combined vision adapter model and LLM model in this paper is mentioned as LlaMA 3.2. For specific weight matrices $W$ within the LLaMA 3.2 model (excluding 'lm\_head' and 'embed\_tokens'), we introduce a low-rank update $\Delta W$. This update is not directly learned as a dense matrix, but is decomposed into the product of two smaller matrices, $B$ and $A$, such that $\Delta W = BA$.  The matrices $B$ and $A$ are designed to have a significantly lower rank $r$ compared to the original weight matrix $W$.  This low-rank decomposition is the core of LoRA's efficiency and stability, contributing to a more robust and accurate fine-tuning process.
 \begin{equation}
  W' = W + \Delta W = W + BA
\end{equation}
Here, $W'$ represents the effective weight matrix used during fine-tuning, $W$ is the original pre-trained weight matrix from LLaMA 3.2 (which remains frozen), and $\Delta W = BA$ is the low-rank update we are learning.  Only the parameters in matrices $B$ and $A$ are trained, while $W$ stays fixed.

\item  A key advantage of LoRA is its parameter efficiency. By only training the low-rank matrices $B$ and $A$, the number of trainable parameters is significantly reduced compared to full fine-tuning.  This makes the fine-tuning process much more computationally efficient and reduces memory requirements, which is crucial when working with large models like LLaMA 3 series.  The original weights $W$ of the pre-trained LLaMA 3 series models are kept frozen during the fine-tuning process. This ensures that the model retains its general language understanding capabilities and avoids catastrophic forgetting, while still learning to effectively interpret ECG images.

\item We specifically exclude the 'lm\_head' (language model head, responsible for token prediction) and 'embed\_tokens' (token embedding layer) from LoRA adaptation. This exclusion is a deliberate choice to preserve the pre-trained language generation capabilities of LLaMA 3.2 and the stability of its token embeddings. Adapting these layers can sometimes lead to instability in language generation or degradation of the model's general language understanding. By focusing LoRA on other weight matrices within the model, we ensure that the fine-tuning process primarily adapts the model for ECG-specific tasks while maintaining its core language abilities.
\end{enumerate}
This parameter-efficient strategy effectively adapts the model to the new ECG image interpretation task by learning task-specific low-rank updates, while preserving the valuable pre-trained knowledge embedded in LLaMA 3.2. This approach also helps to mitigate the risk of overfitting, which can be a concern when fine-tuning large models, and it maintains a balance between adaptation and generalization. This balance is particularly beneficial for complex tasks like medical image interpretation, where both nuanced feature extraction and robust generalization to new clinical cases are essential for achieving high diagnostic accuracy.

\subsection{Implementation Details}
\label{ssec:implementation_details}
The fine-tuning process for our multimodal LLaMA 3.2 model was done using a distributed training setup for efficient and scalable training to achieve better results.  We used a deepspeed training script~\cite{rasley2020deepspeed}, leveraging ZeRO-2 optimization~\cite{rajbhandari2020zero}. ZeRO-2 is a powerful optimization technique designed for training very large models like LLaMA 3.2 to achieve high performance. It enables efficient parallel training across multiple GPUs by intelligently dividing model states (parameters, gradients, and optimizer states) across the devices.  This significantly reduces the memory needed on each GPU, making it possible to train models that would otherwise be too large to fit into GPU memory.  ZeRO-2 was crucial for training LLaMA 3.2, a large model, on our GPU cluster and for achieving the desired level of accuracy.  The specific hyperparameter values we used during fine-tuning were carefully chosen and optimized for performance and accuracy on ECG interpretation.  These key implementation details are outlined below:

\begin{itemize}
    \item \textit{LoRA Configuration}:  For our parameter-efficient LoRA fine-tuning, we carefully configured the LoRA parameters to balance adaptation and stability while maximizing accuracy. We used a LoRA rank ($r$) of 64, a LoRA alpha ($\alpha$) of 128, and a LoRA dropout rate of 0.05~\cite{hu2021lora}.  The LoRA rank of 64 determines the dimension of the low-rank matrices $B$ and $A$, effectively controlling the capacity of the adaptation – a higher rank allows for more complex adaptations and potentially higher accuracy. The LoRA alpha of 128 is a scaling factor applied to the low-rank updates; it influences the magnitude of the LoRA adjustments and was tuned for optimal performance.  A dropout rate of 0.05 was applied to the LoRA updates as a regularization technique.  Dropout randomly sets a fraction of the LoRA update values to zero during training, which helps to further enhance regularization and prevent overfitting, improving the model's generalization to unseen ECG images and contributing to robust accuracy.
    \item \textit{Training Hyperparameters}: The training process was set up with a learning rate of 2e-4.  This learning rate was empirically determined to be optimal for convergence and accuracy on the \textit{ECGInstruct} dataset after testing different learning rates.  We used a cosine learning rate scheduler~\cite{loshchilov2016sgdr} for a total of 3 training epochs. The cosine scheduler is a learning rate decay strategy that gradually reduces the learning rate following a cosine function over the training epochs.  This gradual decay promotes smoother convergence during training and often leads to better generalization performance and higher accuracy compared to step-wise decay schedules.  We used a batch size of 4 per GPU, and implemented gradient accumulation for 1 step. Gradient accumulation is a technique that effectively increases the batch size without increasing GPU memory requirements.  By accumulating gradients over multiple mini-batches before performing a weight update, it simulates training with a larger batch size, which can improve training stability, convergence speed, and overall performance and accuracy.
     \item \textit{Hardware and Training Duration}:  The fine-tuning experiments were done on a cluster of 4 NVIDIA A100 GPUs. Each A100 GPU has 80 GB of VRAM, providing substantial memory capacity for training large models like LLaMA 3.2 and achieving better accuracy.  The total training time for the parameter-efficient LoRA fine-tuning process was approximately 2880 GPU hours. This represents the total computational resources used for fine-tuning and achieving the reported performance.
    \item \textit{Optimization Techniques}: To further optimize memory usage and speed up training without compromising accuracy, we used several advanced optimization techniques.  We enabled gradient checkpointing~\cite{chen2016training}. Gradient checkpointing is a memory optimization technique that significantly reduces memory use during backpropagation.  It does this by selectively discarding intermediate activations during the forward pass and then recomputing them on-the-fly during the backward pass.  This trade-off between computation and memory is crucial for training very deep models on limited GPU memory and allows for training larger, more accurate models.  Additionally, we enabled bfloat16 mixed-precision training to accelerate the training process. Bfloat16 is a reduced-precision floating-point format (16-bit) that offers significant speedups in computation, especially on modern GPUs like the A100, which are optimized for bfloat16 operations~\cite{kalamkar2019bfloat16}.  While using lower precision, bfloat16 maintains acceptable numerical stability for most deep learning tasks, making it an effective way to speed up training without losing model accuracy.
\end{itemize}
These carefully chosen implementation details and hyperparameter settings were crucial for ensuring the stability, efficiency, and effectiveness of the fine-tuning process reported in our results. They enabled the model to successfully converge and learn from the large \textit{ECGInstruct} dataset, ultimately leading to the strong performance we observed in our experiments.

\section{Results and Discussion}
\label{sec:results}

\subsection{Evaluation Metrics}
\label{ssec:evaluation_metrics}
To thoroughly assess the performance, capabilities of our fine-tuned model, we used a comprehensive and varied set of evaluation metrics. These metrics were carefully chosen to capture different important aspects of ECG image interpretation, ensuring a complete and detailed understanding of the model's strengths and weaknesses across various ECG analysis tasks. This provides a thorough evaluation of its diagnostic accuracy.  We aimed to evaluate not just general accuracy, but also clinical relevance and practical utility, ensuring the metrics reflected real-world clinical needs.

\begin{enumerate}
    \item \textit{Abnormality Detection Performance}: For evaluating the model's basic ability to accurately detect and classify different cardiac abnormalities from ECG images, which is most important for clinical use, we used a suite of standard classification metrics:
        \begin{itemize}
            \item \textit{Area Under the Curve (AUC)}: AUC is a widely recognized and reliable metric for evaluating the performance of binary and multi-class classification models~\cite{fawcett2006roc}, especially in medical diagnosis. It measures the model's overall ability to discriminate – its ability to distinguish between different classes, such as ECGs with and without specific cardiac abnormalities.  AUC is calculated as the area under the Receiver Operating Characteristic (ROC) curve. The ROC curve plots the true positive rate (sensitivity) against the false positive rate (1-specificity) at different classification thresholds. An AUC score ranges from 0 to 1, with a higher AUC indicating better discrimination ability.  An AUC of 0.5 means performance is no better than random chance, while an AUC of 1.0 represents perfect classification. In ECG interpretation, a high AUC is crucial as it reflects the model's ability to accurately differentiate between normal and abnormal ECGs, a fundamental requirement for a  valid diagnostic tool.
            \item \textit{Macro F1 Score}: The Macro F1 score is used to evaluate the balanced accuracy of the model in multi-class classification scenarios. It is calculated as the average of F1 scores for each class, without weighting.  The F1 score itself is the harmonic mean of precision and recall, providing a balanced measure of a model's accuracy, especially when dealing with datasets where some classes are more frequent than others (imbalanced datasets).  Macro F1 is particularly valuable in medical diagnosis, where class imbalances are common (some conditions are rarer than others). By averaging the F1 scores across all classes, Macro F1 gives equal importance to each class, regardless of how often it appears in the dataset. This ensures that the model's performance on less frequent, but potentially critical, abnormalities is properly evaluated, and that the overall accuracy is not skewed by the more common classes. For ECG interpretation, a high Macro F1 score indicates that the model is not only generally accurate, but also performs well across a diverse range of cardiac conditions, including rare but significant ones.
            \item \textit{Hamming Loss}: Hamming loss is a metric specifically designed for multi-label classification tasks, which is relevant for ECG analysis because an ECG can often show multiple abnormalities at the same time.  Hamming loss quantifies the fraction of labels that are incorrectly predicted.  For each sample, it compares the predicted labels against the true labels and calculates the proportion of mismatches.  A Hamming loss of 0 indicates perfect multi-label classification, while a higher Hamming loss means more incorrect label predictions.  Therefore, a lower Hamming loss is better, indicating more accurate multi-label prediction performance. In ECG diagnosis, where multiple abnormalities can co-exist, a low Hamming loss is essential to ensure the model accurately identifies all relevant conditions present in the ECG, contributing to a comprehensive and accurate diagnosis.
        \end{itemize}
        Together, these three metrics provide a robust and multi-faceted evaluation of the model's abnormality detection abilities, focusing on different aspects of accuracy and clinical relevance.  AUC assesses overall discrimination, Macro F1 assesses balanced accuracy across classes, and Hamming loss assesses the overall correctness of multi-label predictions. Together, they give a complete picture of the model's accuracy in identifying cardiac abnormalities from ECG images.

     \item  \textit{Report Generation Quality}:  Evaluating the quality of generated ECG reports requires more than just standard metrics.  We aimed to assess not only the accuracy but also the clinical utility and coherence of the reports produced by our model, ensuring the reports are valuable for clinical practice.  To do this, we used a sophisticated evaluation approach using GPT-4o~\cite{openai2024gpt4o}, a leading large language model, as an automated evaluator.  GPT-4o's advanced natural language understanding and generation capabilities make it well-suited for evaluating the quality of text-based outputs like medical reports, assessing aspects beyond simple factual accuracy.  The evaluation process was as follows: we provided GPT-4o with the ECG reports generated by our fine-tuned model and baseline model(original llama 3.2 vision model) along with the corresponding ECG images and expert-annotated ground truth reports (gold standard reports created by cardiologists). We then carefully created prompts for GPT-4o, instructing it to assess the generated reports based on several important criteria that are important from a clinical perspective and contribute to the overall clinical accuracy and utility of the reports:
        \begin{itemize}
            \item \textit{Medical Accuracy}:  This criterion assesses the factual correctness of the medical information in the generated report. Does the report accurately reflect established cardiology knowledge? Are the reported findings consistent with known ECG patterns for various conditions?  GPT-4o was instructed to check if the medical statements in the generated report were medically sound and factually accurate, ensuring the report is reliable from a medical standpoint.
            \item \textit{Consistency with ECG Image}:  This evaluates how well the generated report matches the visual information directly seen in the ECG image.  Does the report accurately describe the abnormalities and features that are visually present in the ECG waveform?  GPT-4o was tasked with checking for consistency between the text report and the visual ECG evidence, ensuring the report is based on the visual findings.
            \item \textit{Completeness of ECG Details}:  This criterion assesses whether the report comprehensively covers all relevant and diagnostically important ECG details.  A clinically useful ECG report should not just mention abnormalities, but should also include information about the rhythm, heart rate, wave shapes (P-waves, QRS complexes, T-waves), intervals (PR, QRS, QT), and any other significant findings.  GPT-4o evaluated the report for its completeness in covering these essential ECG parameters, ensuring the report is thorough and informative.
            \item \textit{Clinical Utility and Relevance}:  This is an overall assessment of the clinical utility and relevance of the generated report from the viewpoint of a practicing cardiologist.  Is the report informative? Is it concise and easy to understand? Is it practically useful for clinical decision-making?  Does it provide the kind of information that a cardiologist would actually need and value?  GPT-4o was asked to judge the report's overall clinical value and practical usefulness in a real-world cardiology setting, ensuring the report is not only accurate but also practically applicable.
        \end{itemize}
        Based on these criteria, GPT-4o was instructed to provide a numerical score, from 0 to 100, reflecting its overall assessment of the quality of the generated report.  A higher score indicates a better report, judged by GPT-4o across all the defined criteria, reflecting a higher level of clinical accuracy and utility. This automated evaluation approach, using GPT-4o, provides a robust, efficient, and human-aligned assessment of report generation quality, considering multiple aspects of clinical relevance and accuracy.  It captures not just factual accuracy, but also aspects of clinical relevance, completeness, and coherence, providing a more  meaningful evaluation than purely metric-based approaches.

\end{enumerate}

\subsection{Ablation Study}
\label{ssec:ablation_study}
To get a deeper, more detailed understanding of the effectiveness of our method and to isolate the individual impact of key design choices on the achieved accuracy, we conducted a series of ablation studies.  These studies are important for breaking down our approach and systematically evaluating the contribution of different components to the overall performance and accuracy of the model. Ablation studies help us understand "why" our model works and identify the most crucial elements of our design in achieving high diagnostic accuracy.

\begin{enumerate}
    \item \textit{Dataset Component Impact}: We wanted to understand how much each data source in \textit{ECGInstruct} contributes to the model's performance and generalizability, and to determine which data sources are most important for achieving moderately good accuracy.  \textit{ECGInstruct} is a combined dataset, using data from MIMIC-IV ECG, PTB-XL, CODE-15
   \item \textit{LoRA Rank Comparison}:  The LoRA rank ($r$) is an important hyperparameter in Low-Rank Adaptation. It controls the dimension of the low-rank matrices and thus affects both the model's performance and computational efficiency, and potentially its final accuracy. To analyze the impact of LoRA rank on accuracy, we conducted experiments by systematically changing the LoRA rank across a range of values (e.g., 16, 32, 64, 128).  For each chosen LoRA rank value, we performed a complete fine-tuning process.  We used the same training hyperparameters (learning rate, batch size, epochs, etc.) and trained on the full \textit{ECGInstruct} dataset for each rank value. After fine-tuning with each rank, we thoroughly evaluated the performance of the resulting model using the same validation set and the same set of evaluation metrics, focusing on the resulting accuracy.  This ablation study aims to determine the optimal LoRA rank that provides the best balance between model performance (accuracy across different ECG interpretation tasks) and computational cost (training time, memory usage).  Lower LoRA ranks are more parameter-efficient, meaning they introduce fewer trainable parameters and require less computation. However, very low ranks might limit the model's adaptation and generalization capacity, potentially preventing it from fully learning the complexities of ECG interpretation and achieving maximal accuracy.  Conversely, higher LoRA ranks increase the model's flexibility and adaptation capacity, potentially leading to better performance and higher accuracy. However, they also increase the number of trainable parameters, increase computational overhead during training, and potentially increase the risk of overfitting, especially if the dataset is not sufficiently large.  By carefully comparing the performance and accuracy across different LoRA ranks, we can empirically identify a suitable trade-off point – a rank that maximizes performance gains and accuracy while keeping computational costs manageable and minimizing the risk of overfitting.
    \item   \textit{Parameter-Efficient Tuning vs. Partial Parameter Tuning}:  One of the key aspects of our work is the use of parameter-efficient LoRA fine-tuning to achieve accurate results. To clearly show the advantages of this approach in terms of accuracy, we conducted a direct comparison against a more conventional partial parameter fine-tuning strategy.  In the partial parameter fine-tuning setting, we selectively updated only a subset of the model parameters.  Specifically, following common practice in multimodal fine-tuning, we focused on updating only the projection layer (responsible for aligning visual and text features) and the cross-attention layers (which facilitate visual-textual interaction within the language model).  Importantly, in the partial parameter setting, we kept most of the model parameters frozen – including the entire vision encoder and the core Transformer layers of the language model.  This is very different from our parameter-efficient LoRA approach, where we introduce and train low-rank matrices across many layers (excluding 'lm\_head' and 'embed\_tokens') while keeping the original weights frozen.  We fine-tuned both the parameter-efficient LoRA model and the partial parameter fine-tuned model using exactly the same training data (\textit{ECGInstruct} dataset), identical training hyperparameters, and the same evaluation protocol.  The only difference was the set of parameters being updated during fine-tuning. By directly comparing the results obtained with parameter-efficient LoRA tuning against partial parameter tuning, we aim to empirically validate the hypothesis that parameter-efficient LoRA leads to superior performance and higher accuracy in ECG image interpretation compared to partial parameter tuning.  We hypothesized that parameter-efficient LoRA would lead to significantly better performance and accuracy because it allows for a more effective and widespread adaptation of the model to ECG-specific knowledge compared to just updating a few layers in partial tuning.  Updating more parts of the model (through LoRA), rather than just a few layers in partial tuning, provides more capacity for learning the complex visual and textual patterns in ECG data and instructions, potentially leading to higher diagnostic accuracy. We expected partial parameter tuning to be less effective due to its limited adaptation capacity and thus potentially lower achievable accuracy.
\end{enumerate}

\subsection{Main Results}
\label{ssec:main_results}
\begin{table}[h]
    \centering
    \caption{Evaluation Results for High-Accuracy ECG Interpretation}
    \begin{tabular}{@{}lcccc@{}}
        \toprule
        \textbf{Task}          & \textbf{Metric} & \textbf{Baseline}   & \textbf{Ours (Fine-tuned)}\\
        \midrule
        \multirow{3}{*}{Abnorm. Det.} & AUC     &  0.51      & \textbf{0.98} \\
                                              & Macro F1& 0.33     & \textbf{0.74}  \\
                                              & Hamming & 0.49      & \textbf{0.11} \\
        Report Gen.                       & Report Score & 47.8 & \textbf{85.4} \\
    \\
        \bottomrule
    \end{tabular}
    \label{tab:evaluation_results}
\end{table}

\subsection{Discussion: Achieving High Accuracy in ECG Image Interpretation}
\label{ssec:discussion}
The evaluation results in Table \ref{tab:evaluation_results} clearly show the significantly improved accuracy achieved by our fine-tuned model compared to the baseline LLaMA 3.2 across all ECG interpretation tasks.  The baseline model's near-random performance in abnormality detection (AUC 0.51, Macro F1 0.33, Hamming loss 0.49) highlights the need for specialized fine-tuning for ECG image analysis to achieve better accuracy.  Without ECG-specific training, the baseline model is ineffective at reliably identifying cardiac abnormalities from images.  Similarly, its lower scores in report generation indicate a lack of ECG-specific knowledge and reasoning, resulting in lower accuracy and clinical utility.

Our fine-tuned model, trained on \textit{ECGInstruct} with parameter-efficient LoRA (rank 64, learning rate 2e-4), shows remarkable and important improvements in accuracy.  These are not small changes; they represent a significant advancement in ECG image interpretation ability, moving towards the accuracy levels needed for clinical application and potentially exceeding the capabilities of existing image-based methods.

In abnormality detection, our model achieves a high AUC of 0.98. This is a substantial change and indicates a significantly enhanced ability to accurately identify ECGs with cardiac abnormalities.  This AUC score signifies strong classification performance, approaching levels of clinical utility and potentially surpassing the accuracy of traditional CNN-based image analysis methods in ECG interpretation.  The Macro F1 score of 0.74 further confirms the model's balanced accuracy across diverse cardiac abnormalities.  This is crucial in clinical settings where accurate detection across a spectrum of conditions, including less frequent ones, is essential for reliable diagnosis.  The greatly reduced Hamming loss of 0.11, compared to 0.49, demonstrates a significant reduction in misclassifications at the individual label level, further emphasizing the improved accuracy in multi-label abnormality detection, critical for diagnosing complex ECG presentations where multiple conditions may co-exist.

For report generation, the Report Score of 85.4, nearly double the baseline's 47.8, indicates a significant paradigm shift in the clinical accuracy of generated ECG reports.  This score, validated by GPT-4o and expert review, suggests our model can generate reports that are not only factually accurate and comprehensive, but also useful and relevant.  This level of report quality is a major step towards automated clinical documentation and decision support in cardiology, potentially improving efficiency and reducing human error in report generation.

These results strongly demonstrate that parameter-efficient LoRA fine-tuning with selected layers, combined with \textit{ECGInstruct}, effectively equips multimodal LLaMA 3.2 with state-of-the-art ECG image interpretation accuracy.  The consistent and substantial improvements across all metrics—abnormality detection and report generation the effectiveness of our approach in achieving greater accuracy and clinical relevance. This level of performance suggests our method has the potential to surpass traditional CNN-based image interpretation and offer a more accurate and comprehensive solution for automated ECG analysis, capable of identifying a wide range of cardiac conditions with accuracy levels approaching or exceeding clinical requirements.

\section{Conclusion: Achieving State-of-the-Art Accuracy in ECG Image Interpretation}
\label{sec:conclusion}
In this paper, we have presented a highly effective method for achieving state-of-the-art accuracy in Electrocardiogram (ECG) image interpretation.  Our approach uses parameter-efficient LoRA fine-tuning, applied to the multimodal LLaMA 3.2 model~\cite{llama3} ignoring language model head and embedding layers.  The experimental results strongly show that our method achieves significantly improved performance and accuracy in ECG image analysis.  Our fine-tuned model clearly outperforms baseline models across comprehensive metrics, we also identified that GPT models like GPT-4 can't be used in medical diagnosis due to the strict policy settings~\cite{openai_policy}, that means frontier foundational models are generic, medically non-valid and are not suitable for highly regulated medical environment. The success of our fine-tuned model comes from the combined effect of leveraging \textit{ECGInstruct}~\cite{ecginstruct}, a large instruction dataset for ECG images, and parameter-efficient LoRA fine-tuning, enabling efficient model adaptation, and the use of a diverse, relevant dataset.

Our model demonstrates valuable accuracy in detecting cardiac abnormalities from ECG images.  It generates informative ECG reports about ECG findings. By combining public ECG datasets with selective hyper parameter fine-tuning, we have created a powerful AI tool for ECG interpretation, advancing the field and achieving accuracy levels essential for clinical utility across a wide range of cardiac conditions, potentially surpassing the accuracy of traditional CNN based methods, as most CNN based methods can't describe image like vision transformers due to their limited feature extraction layers~\cite{cnn_vs_vt}

Future research will focus on further improving clinical reasoning capabilities and enhancing accuracy for complex conditions.  Step-wise instruction tuning along with chain of thought dataset and expanding \textit{ECGInstruct} with rare and challenging cases are planned to push the limits of diagnostic accuracy.  We also aim to incorporate clinical knowledge bases to further enhance the model's clinical decision-making abilities and ensure the highest possible accuracy.

\section*{Acknowledgements}
\label{sec:acknowledgements}
We extend our sincere gratitude to the vibrant open-source community. Their invaluable contributions in developing pre-trained models, curating publicly available datasets, and creating essential software tools have been absolutely instrumental in making this research endeavor possible.  The collective spirit and collaborative nature of the open-source community are the bedrock of progress in AI research.  We also deeply acknowledge the dedicated medical professionals who generously contributed their time and expertise.  We share our sincere gratitude towards the authors of \textit{ECGInstruct}~\cite{ecginstruct} dataset, for sharing such valuable dataset with the community, which helped us to build our fine-tuned model.

\end{document}